# Nonlinear Quality of Life Index

**ANDREI ZINOVYEV**
*Institut Curie, U900 INSERM/Curie/Mines Paritech, rue d'Ulm 26*
*Paris, 75005, France*
*E-mail: andrei.zinovyev@curie.fr*

**ALEXANDER N. GORBAN**
*Department of Mathematics, University of Leicester, University Road*
*Leicester LE1 7RH, UK*
*E-mail: ag153@le.ac.uk*

**Abstract.** We present details of the analysis of the nonlinear quality of life index for 171 countries. This index is based on four indicators: GDP per capita by Purchasing Power Parities, Life expectancy at birth, Infant mortality rate, and Tuberculosis incidence. We analyze the structure of the data in order to find the optimal and independent on expert's opinion way to map several numerical indicators from a multidimensional space onto the one-dimensional space of the quality of life. In the 4D space we found a principal curve that goes "through the middle" of the dataset and project the data points on this curve. The order along this principal curve gives us the ranking of countries.

The measurement of the quality of life is very important for economic and social assessment and also for public policy, social legislation, and community programs. "There is a strong need for a systematic exploration of the content, reach, and relevance of the concept of the quality of life, and ways of making it concrete and usable" [1].

Many of the existing indices of quality of life (for example, The Economist Intelligence Unit's quality-of-life index and The Life Quality Index, LQI) are not free from certain problems and biases. For example, LQI uses a parameter K which cannot be justified from data and should be used *ad hoc* for each group of countries, thus, already introducing bias in the estimation. From the other hand, The Economist Intelligence Unit's index is based on expert-based surveys of the "life satisfaction" that are inevitably biased by psychological and linguistic intercultural differences. Probably, as a result of such biases, relative rankings of some countries are counter-intuitive (for example, Russia is ranked below Turkmenistan and Kyrgyzstan).

In our paper [2] we attempted to analyse the structure of the data that are commonly used for quantifying the quality of life, in order to find the optimal and independent on expert's opinion way to map several numerical indicators from a multidimensional space onto the one-dimensional space of the quality of life. In this paper we present an additional case study using only publicly available data taken from GAPMINDER statistical data compilation [3].

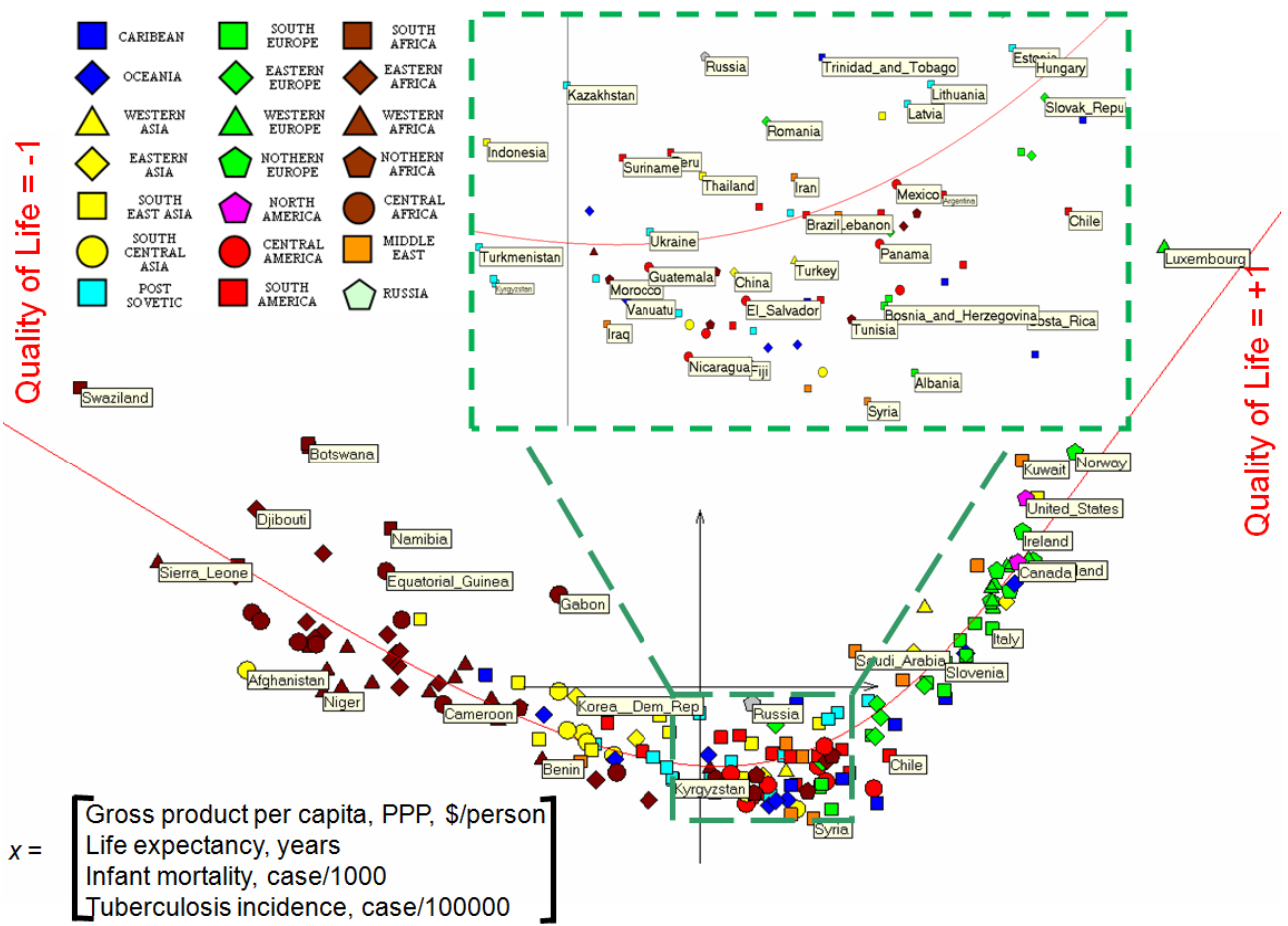

Fig. 1. Approximation of data used for constructing the quality of life index. Each of 171 points represents a country i 4-dimensional space formed by the values of 4 indicators (gross productper capita, life expectancy, infant mortality, tuberculosis incidence). Different forms and colors correspond to various geographical locations. Red line represents the principal curve, approximating the dataset. Three arrows show the basis of principal components. The best linear index (first principal component) explains 75% of variation, while the non-linear index (principal curve) explains 86% of variation.

For illustrating the principle of the index construction, it was decided to use four GAPMINDER indicators for 2005 year:

1. Gross Domestic Product per capita by Purchasing Power Parities, $ per person
2. Life expectancy at birth, years
3. Infant Mortality Rate (per 1000 born)
4. Infectious Tuberculosis, new cases per 100,000 of population, estimated.

The details of computing and collecting these values are available from GAPMINDER web-site [3].

The advantage of using these four indicators is also in that they are well-defined and assessed for the vast majority of modern states. Of course, these indicators are widely used, for example, in the most celebrated Human Development Index [4-6] but we use an unsupervised nonlinear Principal Component Analysis [8,9] instead of direct construction of index by experts, in order to reduce arbitrariness of these constructions. We avoid using more sophisticated indicators like Divorce rate or Rate of church attendance because the interpretation of their impact onto quality of life is not so obvious. In that sense, we are closer to the Physical Quality of Life Index [7]. We use the Gross product together with direct physical (physiological) indicators because it reflects the wealth of the countries and the spectrum of the possibilities for people to build their life.

The data were rescaled to z-values (such that the variance for each variable would equal unity). Exploratory analysis performed with use of Principal Component Analysis, produced the distribution shown at the Fig. 1. This distribution allows us to make a conclusion that *no linear function exist that can equally well serve for reducing the dimension of this space from 4 to 1*. The distribution is intrinsically curved, hence, any linear mapping will inevitably give strong distortions in one or other region of dataspace.

There is a simple reason for this in the data set structure. Observing Fig. 1, it is easy to realize that all countries can be roughly separated in two groups. The first group consists of very wealthy countries, mostly from Western and Nothern Europe, USA, Australia and some others (right branch of the distribution on the Fig.1). It happens that the most of variation among these countries can be attributed to the gross product per capita feature while others are approximately equal for them and do not contribute significantly to the variance.

The second group (left branch of the distribution on Fig. 1) consists of very poor countries (mostly, African), which are "equally poor" in terms of the gross product per capita but can be very different in terms of their problems (lower of higher life expectancy, level of infectious diseases, conditions of the state health system), for a number of reasons (wars, difference in internal politics, etc.).

However, this classification in two classes is not perfect: in fact, there are many countries which are localized in the non-linear junction between these two branches of the 4-dimensional distribution. This intermediate group includes, for example, most of the Post-Soviet countries.

We have constructed a principal curve approximating the curved data distribution in the four-dimensional space (red line on Fig. 1), using the Elastic Map approach. The advantage of using the principal curve instead of first principal component was significant in term of Mean-Squared Error (86% vs 76% of explained variance).

The projection of a country onto the principal curve served as a value of the Non-linear index of Quality of Life (NQL) in the range from -1 (left end of the curve) to +1 (right end of the curve). The order of these projections serves for ranking of countries (Table 1).

Importantly, using the non-linear index for non-linear one-dimensional dimension reduction can significantly change the absolute and relative rankings for many countries, including Russia. Thus, for the index, constructed using the first principal component, Russia in 2005 is ranked the 88[th] (with the most prosperous Luxembourg at the 1[st] place), after, for example, Egypt (86[th]). For non-linear index, Russia is ranked the 79[st] before Egypt (87[th]).

In our approach, the value of GDP per capita is important but does not dominate over the physical quality of life indicators. For example, Equatorial Guinea has big GDP per capita (higher than Iran, for example), nevertheless, it is ranked rather low, as #151 among 171 countries. Russia (#79) has higher GDP per capita than 23 preceding countries (## 56-78) but worse physical quality of life (lower Life expectancy and higher Tuberculosis incidence) moved Russia to #79.

The method of Principal manifolds is presented in detail in the book of reviews [10].

**Table 1.** 2005 nonlinear quality of life index: Data, countries rating and ranking

| Country | Gross Domestic Product per capita by Purchasing Power Parities | Life expectancy at birth, years | Tuberculosis, new cases per 100,000 of population | Infant Mortality Rate (per 1000 born) | Non-linear Quality of Life Index | Ranking |
|---|---|---|---|---|---|---|
| Luxembourg | 70014 | 79.56 | 6 | 4 | 0.892 | 1 |
| Norway | 47551 | 80.29 | 3 | 3 | 0.647 | 2 |
| Kuwait | 44947 | 77.258 | 11 | 10 | 0.608 | 3 |
| Singapore | 41479 | 79.627 | 12 | 2 | 0.578 | 4 |
| United States | 41674 | 77.93 | 2 | 7 | 0.575 | 5 |
| Ireland | 38058 | 79.4 | 6 | 4 | 0.539 | 6 |
| Iceland | 35630 | 81.43 | 2 | 2 | 0.519 | 7 |
| Switzerland | 35520 | 81.45 | 3 | 4 | 0.517 | 8 |
| Canada | 35078 | 80.36 | 2 | 5 | 0.508 | 9 |
| Netherlands | 34724 | 79.65 | 3 | 4 | 0.503 | 10 |
| Austria | 34108 | 79.64 | 6 | 4 | 0.495 | 11 |
| Denmark | 33626 | 78.32 | 4 | 4 | 0.486 | 12 |
| Australia | 32798 | 81.44 | 3 | 5 | 0.486 | 13 |
| United Arab Emirates | 33487 | 77.055 | 7 | 8 | 0.479 | 14 |
| Sweden | 31995 | 80.75 | 3 | 3 | 0.475 | 15 |
| Belgium | 32077 | 79.21 | 6 | 4 | 0.471 | 16 |
| United Kingdom | 31580 | 79.3 | 6 | 5 | 0.465 | 17 |
| Japan | 30290 | 82.27 | 11 | 3 | 0.460 | 18 |
| Germany | 30496 | 79.48 | 3 | 4 | 0.454 | 19 |
| Finland | 30469 | 79.09 | 3 | 3 | 0.453 | 20 |
| France | 29644 | 80.47 | 6 | 4 | 0.447 | 21 |
| Italy | 27750 | 81.18 | 3 | 4 | 0.428 | 22 |
| Spain | 27270 | 80.28 | 13 | 4 | 0.418 | 23 |
| Bahrain | 27236 | 75.234 | 19 | 9 | 0.395 | 24 |
| Greece | 25520 | 78.666 | 8 | 4 | 0.393 | 25 |
| New Zealand | 24554 | 79.799 | 4 | 5 | 0.386 | 26 |
| Cyprus | 24473 | 79.401 | 2 | 4 | 0.384 | 27 |
| Israel | 23846 | 80.02 | 3 | 4 | 0.380 | 28 |
| Slovenia | 23004 | 77.6 | 6 | 3 | 0.360 | 29 |

| Country | | | | | | |
|---|---:|---:|---:|---:|---:|---:|
| Malta | 20410 | 79.245 | 3 | 5 | 0.336 | 30 |
| Korea, Rep. | 21342 | 78.585 | 38 | 5 | 0.336 | 31 |
| Puerto Rico | 19725 | 78.401 | 2 | 0 | 0.329 | 32 |
| Portugal | 20006 | 78.2 | 15 | 4 | 0.324 | 33 |
| Czech Rep. | 20281 | 76.22 | 5 | 3 | 0.321 | 34 |
| Oman | 20334 | 74.951 | 5 | 10 | 0.309 | 35 |
| Saudi Arabia | 21220 | 72.209 | 19 | 21 | 0.288 | 36 |
| Barbados | 15837 | 76.531 | 2 | 11 | 0.263 | 37 |
| Hungary | 17014 | 72.99 | 10 | 6 | 0.259 | 38 |
| Slovak Republic | 15881 | 74.2 | 8 | 7 | 0.252 | 39 |
| Estonia | 16654 | 72.96 | 18 | 6 | 0.251 | 40 |
| Chile | 12262 | 77.93 | 8 | 8 | 0.233 | 41 |
| Poland | 13573 | 75.15 | 12 | 6 | 0.231 | 42 |
| Croatia | 13232 | 75.53 | 18 | 6 | 0.227 | 43 |
| Lithuania | 14085 | 71.7 | 28 | 7 | 0.203 | 44 |
| Costa Rica | 8661 | 78.481 | 6 | 11 | 0.195 | 45 |
| Cuba | 7407 | 77.928 | 4 | 5 | 0.191 | 46 |
| Latvia | 13218 | 71.24 | 28 | 8 | 0.187 | 47 |
| Argentina | 11063 | 74.773 | 18 | 14 | 0.184 | 48 |
| Uruguay | 9266 | 75.714 | 12 | 12 | 0.180 | 49 |
| Libya | 10804 | 73.363 | 8 | 18 | 0.170 | 50 |
| Grenada | 9128 | 74.894 | 2 | 17 | 0.170 | 51 |
| Malaysia | 11466 | 73.657 | 46 | 10 | 0.169 | 52 |
| Mexico | 11317 | 75.488 | 12 | 30 | 0.168 | 53 |
| Mauritius | 10155 | 72.035 | 10 | 13 | 0.160 | 54 |
| Trinidad and Tobago | 15352 | 68.745 | 5 | 32 | 0.154 | 55 |
| Bulgaria | 9353 | 72.53 | 18 | 12 | 0.153 | 56 |
| Venezuela | 9876 | 73.249 | 18 | 18 | 0.152 | 57 |
| Belize | 7290 | 75.437 | 21 | 15 | 0.147 | 58 |
| Panama | 8399 | 75.171 | 25 | 19 | 0.145 | 59 |
| Albania | 5369 | 76.209 | 9 | 16 | 0.142 | 60 |
| Macedonia, FYR | 7393 | 73.82 | 13 | 15 | 0.142 | 61 |
| Bosnia and Herzegovina | 6506 | 74.821 | 23 | 13 | 0.137 | 62 |
| Lebanon | 10212 | 71.5 | 5 | 27 | 0.131 | 63 |
| Tunisia | 6461 | 73.462 | 11 | 20 | 0.121 | 64 |
| Syria | 4059 | 73.645 | 8 | 13 | 0.119 | 65 |
| Brazil | 8596 | 71.669 | 27 | 20 | 0.113 | 66 |
| Iran | 10692 | 70.618 | 11 | 31 | 0.112 | 67 |
| Colombia | 6306 | 72.25 | 20 | 17 | 0.109 | 68 |
| Belarus | 8541 | 68.84 | 27 | 12 | 0.105 | 69 |
| Romania | 9374 | 72.04 | 60 | 16 | 0.104 | 70 |
| Sri Lanka | 3481 | 73.663 | 27 | 12 | 0.103 | 71 |
| Jamaica | 7132 | 71.278 | 3 | 26 | 0.103 | 72 |
| Turkey | 7786 | 71.396 | 13 | 26 | 0.101 | 73 |
| Tonga | 5135 | 71.428 | 11 | 20 | 0.094 | 74 |

| Country | | | | | | |
|---|---|---|---|---|---|---|
| Jordan | 4294 | 71.919 | 2 | 22 | 0.094 | 75 |
| Ecuador | 6533 | 74.674 | 58 | 22 | 0.090 | 76 |
| Samoa | 4872 | 70.807 | 9 | 24 | 0.079 | 77 |
| Armenia | 3903 | 73.129 | 32 | 23 | 0.074 | 78 |
| Russia | 11861 | 65.33 | 46 | 15 | 0.073 | 79 |
| El Salvador | 5403 | 70.735 | 23 | 23 | 0.072 | 80 |
| China | 4909 | 72.555 | 45 | 21 | 0.070 | 81 |
| Fiji | 4209 | 68.31 | 10 | 16 | 0.069 | 82 |
| Paraguay | 3900 | 71.286 | 32 | 20 | 0.065 | 83 |
| Thailand | 6869 | 68.422 | 62 | 8 | 0.063 | 84 |
| Algeria | 6011 | 71.699 | 24 | 34 | 0.058 | 85 |
| Honduras | 3266 | 71.515 | 34 | 24 | 0.052 | 86 |
| Egypt | 5049 | 69.542 | 10 | 31 | 0.051 | 87 |
| Peru | 6466 | 72.463 | 76 | 23 | 0.045 | 88 |
| Vietnam | 2142 | 73.828 | 77 | 16 | 0.044 | 89 |
| Maldives | 4017 | 69.999 | 24 | 28 | 0.043 | 90 |
| Nicaragua | 2611 | 71.917 | 29 | 30 | 0.041 | 91 |
| Georgia | 3505 | 71.55 | 37 | 29 | 0.038 | 92 |
| Ukraine | 5583 | 67.31 | 44 | 20 | 0.028 | 93 |
| Guatemala | 4897 | 69.641 | 35 | 32 | 0.023 | 94 |
| Suriname | 7234 | 68.425 | 53 | 30 | 0.011 | 95 |
| Vanuatu | 3477 | 69.257 | 37 | 31 | 0.011 | 96 |
| Moldova | 2362 | 67.923 | 63 | 17 | 0.002 | 97 |
| Morocco | 3547 | 70.443 | 44 | 36 | 0.002 | 98 |
| Cape Verde | 2831 | 70.52 | 69 | 26 | -0.001 | 99 |
| Iraq | 3200 | 68.495 | 25 | 37 | -0.002 | 100 |
| Micronesia, Fed. Sts. | 5508 | 68.029 | 47 | 34 | -0.009 | 101 |
| Kazakhstan | 8699 | 64.774 | 61 | 27 | -0.023 | 102 |
| Kyrgyzstan | 1728 | 67.12 | 55 | 37 | -0.055 | 103 |
| Indonesia | 3234 | 69.697 | 107 | 28 | -0.056 | 104 |
| Uzbekistan | 1975 | 67.386 | 53 | 40 | -0.057 | 105 |
| Philippines | 2932 | 71.067 | 135 | 25 | -0.068 | 106 |
| Turkmenistan | 4247 | 64.372 | 31 | 47 | -0.073 | 107 |
| Comoros | 1063 | 64.08 | 21 | 53 | -0.091 | 108 |
| Azerbaijan | 4648 | 69.391 | 34 | 74 | -0.102 | 109 |
| Guyana | 3232 | 65.467 | 64 | 47 | -0.104 | 110 |
| Mongolia | 2643 | 65.5 | 92 | 36 | -0.108 | 111 |
| Nepal | 1081 | 65.251 | 81 | 49 | -0.137 | 112 |
| Sao Tome and Principe | 1460 | 64.924 | 47 | 63 | -0.137 | 113 |
| Solomon Islands | 1712 | 64.812 | 64 | 56 | -0.139 | 114 |
| Bolivia | 3618 | 64.69 | 90 | 52 | -0.156 | 115 |
| Eritrea | 685 | 58.37 | 40 | 50 | -0.157 | 116 |
| Tajikistan | 1413 | 65.607 | 86 | 59 | -0.166 | 117 |
| Laos | 1811 | 63.61 | 69 | 62 | -0.171 | 118 |
| India | 2126 | 62.738 | 75 | 59 | -0.178 | 119 |

| Country | | | | | |
|---|---|---|---|---|---|
| Bangladesh | 1268 | 64.582 | 102 | 54 | -0.181 | 120 |
| Yemen, Rep. | 2276 | 61.507 | 37 | 76 | -0.188 | 121 |
| Korea, Dem. Rep. | 1597 | 66.821 | 155 | 42 | -0.193 | 122 |
| Pakistan | 2396 | 65.553 | 81 | 79 | -0.210 | 123 |
| Bhutan | 3694 | 64.815 | 121 | 65 | -0.229 | 124 |
| Benin | 1390 | 60.172 | 39 | 89 | -0.232 | 125 |
| Papua New Guinea | 1747 | 59.979 | 109 | 55 | -0.235 | 126 |
| Myanmar | 1169 | 60.592 | 75 | 75 | -0.237 | 127 |
| Sudan | 2249 | 57.3 | 103 | 62 | -0.268 | 128 |
| Timor-Leste | 2203 | 59.69 | 145 | 52 | -0.276 | 129 |
| Gabon | 12742 | 59.461 | 133 | 60 | -0.278 | 130 |
| Ghana | 1225 | 56.539 | 89 | 75 | -0.287 | 131 |
| Madagascar | 988 | 58.932 | 108 | 74 | -0.288 | 132 |
| Senegal | 1676 | 54.919 | 115 | 61 | -0.301 | 133 |
| Haiti | 1175 | 60.414 | 161 | 63 | -0.316 | 134 |
| Togo | 888 | 61.358 | 174 | 71 | -0.344 | 135 |
| Mauritania | 1691 | 56.417 | 134 | 78 | -0.349 | 136 |
| Cameroon | 1995 | 50.621 | 82 | 87 | -0.349 | 137 |
| Tanzania | 1018 | 53.651 | 131 | 76 | -0.363 | 138 |
| Guinea | 946 | 56.208 | 112 | 100 | -0.369 | 139 |
| Somalia | 933 | 49.542 | 110 | 91 | -0.397 | 140 |
| Malawi | 691 | 51.057 | 149 | 79 | -0.411 | 141 |
| Kenya | 1359 | 52.456 | 162 | 79 | -0.417 | 142 |
| Cambodia | 1453 | 59.436 | 222 | 67 | -0.419 | 143 |
| Uganda | 991 | 50.251 | 151 | 79 | -0.420 | 144 |
| Ethiopia | 591 | 53.762 | 171 | 80 | -0.421 | 145 |
| Burkina Faso | 1140 | 52.088 | 101 | 121 | -0.422 | 146 |
| Congo, Rep. | 3621 | 53.28 | 178 | 79 | -0.433 | 147 |
| Cote d'Ivoire | 1575 | 55.939 | 181 | 91 | -0.440 | 148 |
| Guinea-Bissau | 569 | 46.949 | 91 | 121 | -0.444 | 149 |
| Niger | 613 | 49.668 | 74 | 150 | -0.453 | 150 |
| Equatorial Guinea | 11999 | 49.29 | 113 | 123 | -0.460 | 151 |
| Nigeria | 1892 | 47.27 | 138 | 100 | -0.462 | 152 |
| Liberia | 383 | 57.04 | 115 | 157 | -0.464 | 153 |
| Chad | 1749 | 48.476 | 133 | 124 | -0.486 | 154 |
| Mali | 1027 | 47.354 | 136 | 120 | -0.491 | 155 |
| Rwanda | 813 | 48.422 | 171 | 100 | -0.496 | 156 |
| Burundi | 420 | 49.149 | 165 | 109 | -0.497 | 157 |
| Central African Republic | 675 | 46.16 | 148 | 115 | -0.505 | 158 |
| Afghanistan | 874 | 42.88 | 76 | 165 | -0.512 | 159 |
| Mozambique | 743 | 47.658 | 183 | 100 | -0.516 | 160 |
| Namibia | 4547 | 58.609 | 310 | 46 | -0.528 | 161 |
| Angola | 3533 | 45.523 | 119 | 154 | -0.533 | 162 |
| Congo, Dem. Rep. | 330 | 47.629 | 183 | 129 | -0.561 | 163 |
| Zambia | 1175 | 42.883 | 224 | 102 | -0.606 | 164 |

| | | | | | |
|---|---:|---:|---:|---:|---:|
| Lesotho | 1415 | 44.84 | 239 | 102 | -0.614 | 165 |
| Botswana | 12057 | 50.921 | 294 | 87 | -0.633 | 166 |
| South Africa | 8477 | 51.803 | 349 | 55 | -0.652 | 167 |
| Djibouti | 1964 | 54.456 | 330 | 88 | -0.655 | 168 |
| Sierra Leone | 790 | 46.365 | 219 | 160 | -0.664 | 169 |
| Zimbabwe | 538 | 41.681 | 311 | 68 | -0.680 | 170 |
| Swaziland | 4384 | 44.99 | 422 | 110 | -0.876 | 171 |

## Conclusion

Projection onto the principal curve provides a solution to the classical problem of unsupervised ranking of objects. It allows us to find the independent on expert's opinion way to project several numerical indicators from a multidimensional space onto the one-dimensional space of the index values. This projection is, in some sense, optimal and preserves as much information as possible. For computation we used *ViDaExpert*, a tool for visualization and analysis of multidimensional vectorial data [11].